\documentclass{article}
\usepackage{spconf,amsmath,graphicx}
\usepackage{xcolor}
\usepackage{multirow}
\usepackage{amsmath}
\usepackage[linesnumbered,ruled]{algorithm2e}
\usepackage{float}
\usepackage[numbers,sort&compress,square]{natbib}
\def\rset{{\rm I\!\!\!\;R}}

\def\argmax{\mathop{\rm argmax}}

\title{Advancing RNN Transducer Technology for Speech Recognition}
\name{George Saon, Zolt{\'a}n T{\"u}ske, Daniel Bolanos and Brian Kingsbury}
\address{IBM Research AI, Yorktown Heights, USA}
\begin{document}
\ninept
\maketitle
\begin{abstract}
  We investigate a set of techniques for RNN Transducers (RNN-Ts) that
  were instrumental in lowering the word error rate on three different
  tasks (Switchboard 300 hours, conversational Spanish 780 hours and
  conversational Italian 900 hours). The techniques pertain to
  architectural changes, speaker adaptation, language model fusion,
  model combination and general training recipe. First, we introduce a
  novel multiplicative integration of the encoder and prediction
  network vectors in the joint network (as opposed to additive).
  Second, we discuss the applicability of i-vector speaker adaptation
  to RNN-Ts in conjunction with data perturbation. Third, we explore
  the effectiveness of the recently proposed density ratio language
  model fusion for these tasks.  Last but not least, we describe the
  other components of our training recipe and their effect on
  recognition performance. We report a 5.9\% and 12.5\% word error
  rate on the Switchboard and CallHome test sets of the NIST Hub5 2000
  evaluation and a 12.7\% WER on the Mozilla CommonVoice Italian
  test set.

\end{abstract}
\begin{keywords}
End-to-end ASR, recurrent neural network transducer, multiplicative integration
\end{keywords}

\section{Introduction}

End-to-end approaches directly map an acoustic feature sequence to a
sequence of characters or even words without any conditional
independence assumptions. Compared to traditional approaches which
integrate various knowledge sources in a complex search algorithm,
end-to-end methods resulted in a dramatic simplification of both
training and decoding pipelines. This led to a rapidly evolving
research landscape in end-to-end modeling for ASR with Recurrent
Neural Network Transducers (RNN-T)~\cite{graves12} and attention-based
models~\cite{bahdanau16,chan16} being the most prominent examples. Attention
based models are excellent at handling non-monotonic alignment
problems such as translation~\cite{wu16trans}, whereas RNN-Ts are an ideal match for the
left-to-right nature of
speech~\cite{graves13,rao17,battenberg17,wang18,he19,li19,tripathi19,jain19,mcdermott19,variani20,zeyer20,chiu20,saon20}.

Nowadays, end-to-end models can reach unprecedented levels of speech
recognition performance in spite of, or maybe {\em because of}, the
significantly simpler implementations. It has been shown that, given
enough training data, end-to-end models are clearly able to outperform
traditional approaches~\cite{chiu18}. Nevertheless, data sparsity and
overfitting are inherent problems for any direct sequence-to-sequence
model and various approaches have been proposed to introduce useful
variances and mitigate these
issues~\cite{ko15,saon19a,park19,wan13,wang18switchout}.


After a short overview of RNN-T sequence modeling
(section~\ref{model}), section~\ref{multiplicative} investigates an
architectural change and section~\ref{recipe} presents efficient
training and decoding recipes for such models. The proposed techniques
are then evaluated on three different languages (section~\ref{experiments}),
before conclusions are drawn (section~\ref{conclusion}). As will be
shown, the consistent application of these methods will result in
remarkable end-to-end model performance, even if only a few hundred
hours of training data are available.

\section{RNN-T model description}
\label{model}
Borrowing some notations from~\cite{graves12}, RNN-Ts model the conditional
distribution $p({\bf y}|{\bf x})$ of an output sequence ${\bf
  y}=(y_1,\ldots,y_U)\in{\cal Y}^*$ of length $U$ given an input
sequence ${\bf x}=(x_1,\ldots,x_T)\in{\cal X}^*$ of length $T$. The
elements of ${\bf x}$ are typically continuous multidimensional
vectors whereas the elements of ${\bf y}$ belong to an output space
which is typically discrete. $p({\bf y}|{\bf x})$ is expressed as a
sum over all possible alignments ${\bf a}=(a_1,\ldots,a_{T+U})$ that
are consistent with ${\bf y}$:

\begin{equation}
p({\bf y}|{\bf x})=\sum_{{\bf a}\in{\cal B}^{-1}({\bf y})}p({\bf a}|{\bf x})
\label{ll}
\end{equation}
~~\\
The elements of ${\bf a}$ belong to the augmented vocabulary
$\overline{{\cal Y}}={\cal Y}\cup\{\phi\}$ where $\phi$ (called BLANK)
denotes the null output. The mapping ${\cal B}:\overline{{\cal
    Y}}^*\to{\cal Y}^*$ is defined by ${\cal B}({\bf a})={\bf y}$. For
example, if ${\bf y}=(C,A,T)$ and ${\bf x}=(x_1,x_2,x_3,x_4)$ valid
alignments include $(\phi,C,\phi,A,\phi,T,\phi)$,
$(\phi,\phi,\phi,\phi,C,A,T)$, $(C,A,T,\phi,\phi,\phi,\phi)$, etc.
Furthermore, $p({\bf a}|{\bf x})$ can be factorized as follows:

\begin{align}
\nonumber p({\bf a}|{\bf x})= p({\bf a}|{\bf h}) \stackrel{\Delta}{=} \prod_{i=1}^{T+U} p(a_i|h_{t_i},{\cal B}(a_1,\ldots,a_{i-1}))=\\
\prod_{i=1}^{T+U} p(a_i|h_{t_i},y_0,\ldots,y_{u_{i-1}})=\prod_{i=1}^{T+U} p(a_i|h_{t_i},g_{u_i})
\label{factor}
\end{align}
~~\\
where: ${\bf h}=(h_1,\ldots,h_T)=Encoder({\bf x})$ is an embedding of the input sequence computed by an {\em encoder network}, ${\bf
  g}=(g_1,\ldots,g_U)$ is an embedding of the output sequence computed
by a {\em prediction network} via the recursion
$g_u=Prediction(g_{u-1},y_{u-1})$ (with the convention $g_0={\bf 0},y_0=\phi$), and $p(a_i|h_{t_i},g_{u_i})$ is the predictive output
distribution over $\overline{\cal Y}$ computed by a {\em joint network} which is commonly implemented as:

\begin{equation}
p(\cdot|h_t,g_u)={\rm softmax}[{\bf W}^{out}{\rm tanh}({\bf W}^{enc}h_t+{\bf W}^{pred}g_u+b)]
\label{pred}
\end{equation}
~~\\
${\bf W}^{enc}\in\rset^{J\times E}$, ${\bf W}^{pred}\in\rset^{J\times
  P}$ are linear projections that map $h_t\in\rset^{E}$ and
$g_u\in\rset^{P}$ to a joint subspace which, after addition and {\rm
  tanh}, is mapped to the output space via ${\bf
  W}^{out}\in\rset^{|\overline{{\cal Y}}|\times J}$. RNN-Ts are
typically trained to a minimize the negative log-likelihood (NLL) loss
$-\log p({\bf y}|{\bf x})$. From~(\ref{ll}), calculating the
sum by direct enumeration is intractable because the number of all
possible alignments of length $T+U$ is $\left|{\cal B}^{-1}({\bf y})\right|={T+U\choose U}=\frac{(T+U)!}{T!U!}\ge \left(1+\frac{T}{U}\right)^U$
where ${n\choose k}$ denotes the binomial coefficient $n$ {\em choose}
$k$. Luckily, the factorization in~(\ref{factor}) allows for an
efficient forward-backward algorithm with $T\times U$ complexity for
both loss and gradient computation~\cite{graves12,bagby18}.

\section{Multiplicative integration of encoder and prediction network outputs}
\label{multiplicative}
Here, we discuss a simple change to the joint network equation~(\ref{pred}) and its implications:

\begin{equation}
p(\cdot|h_t,g_u)={\rm softmax}[{\bf W}^{out}{\rm tanh}({\bf W}^{enc}h_t\odot{\bf W}^{pred}g_u+b)]
\end{equation}
~~\\
where $\odot$ denotes elementwise multiplication (or Hadamard
product). This modification was inspired by the work of~\cite{wu16}
where the authors use multiplicative integration (MI) in the context
of a recurrent neural network for fusing the information from the
hidden state vector and the input vector. The advantages of MI over
additive integration for fusing different information sources have
also been mentioned in~\cite{jayakumar19} and
are summarized below.\\
{\bf Higher-order interactions} MI allows for second-order
interactions between the elements of $h_t$ and $g_u$ which facilitates
modeling of more complex dependencies between the acoustic and LM
embeddings. Importantly, the shift to second order is achieved with no
increase in the number of parameters or computational complexity. In
theory, feedforward neural nets with sufficient capacity and training
data are universal function approximators; therefore, a joint network
with $(h_t,g_u)$ inputs and multiple layers should be optimal. However
the memory requirements are prohibitive because the input tensor size
for such a network is $N\times T\times U\times (E+P)$ and the memory
for the output tensors is $N\times T\times U\times H\times L$ where
$N$ is the batchsize, $H$ is the number of hidden units and $L$ is the
number of layers. Therefore, for pure training efficiency reasons,
we are
constrained to use shallow (single layer) joint networks, in which case the functional form of the layer becomes important.\\
{\bf Generalizes additive integration} Indeed, when used in
conjunction with biases, the additive terms appear in the expansion
\begin{align}
\nonumber ({\bf W}^{enc}h_t + b^{enc})\odot ({\bf W}^{pred}g_u+b^{pred})={\bf W}^{enc}h_t\odot{\bf W}^{pred}g_u\\
+b^{enc}\odot{\bf W}^{pred}g_u+b^{pred}\odot{\bf W}^{enc}h_t+b^{enc}\odot b^{pred}
\end{align}
~~\\
{\bf Scaling and gating effect} By multiplying $\tilde{h}_t:={\bf
  W}^{enc}h_t$ with $\tilde{g}_u:={\bf W}^{pred}g_u$, one embedding
has a scaling effect on the other embedding. In particular, unlike for
additive interactions, in MI the gradient with respect to one
component is gated by the other component and vice versa. Concretely,
let us denote by ${\cal L}({\bf x}, {\bf y};\theta)=-\log p({\bf
  y}|{\bf x};\theta)$ the NLL loss for one utterance for an RNN-T with
parameters $\theta$. The partial derivatives of the loss with respect
to $\tilde{h}_t$ and $\tilde{g}_u$ are:

\begin{align}
\nonumber\frac{\partial{\cal L}}{\partial \tilde{h}_t} &=\tilde{g}_u\odot\frac{\partial{\cal L}}{\partial (\tilde{h}_t\odot \tilde{g}_u)}\\
\frac{\partial{\cal L}}{\partial \tilde{g}_u} &=\tilde{h}_t\odot\frac{\partial{\cal L}}{\partial (\tilde{h}_t\odot \tilde{g}_u)}
\end{align}

These arguments suggest that multiplicative integration is a good
candidate for fusing the two different information streams coming from
the encoder and the prediction network. We expect multiplicative
RNN-Ts to be, if not better, then at least complementary to additive
RNN-Ts which should be beneficial for model combination.

\section{Training and decoding recipe}
\label{recipe}

In this section, we discuss some of the techniques that are part of our RNN-T
training and decoding recipe. While none of the techniques presented
here are novel, the goal is to show their effectiveness for RNN-Ts in
particular. The techniques can be roughly grouped into: (i) data
augmentation/perturbation and model regularization, (ii) speaker
adaptation and (iii) external language model fusion.\\
{\bf Speed and tempo perturbation}~\cite{ko15} changes the rate of speech in the interval $[0.9,1.1]$ with or without altering the pitch or timbre of the speaker. This technique generates additional replicas of the training data depending on the number of speed and tempo perturbation values.\\
{\bf Sequence noise injection}~\cite{saon19a} adds, with a given probability, the downscaled spectra of randomly selected training utterances to the spectrum of the current training utterance. This technique does not increase the amount of training data per epoch.\\
{\bf SpecAugment}~\cite{park19} masks the spectrum of a training utterance with a random number of blocks of random size in both time and frequency.\\
{\bf DropConnect}~\cite{wan13} zeros out entries randomly in the LSTM hidden-to-hidden transition matrices.\\
{\bf Switchout}~\cite{wang18switchout} randomly replaces labels in the output sequence with labels drawn uniformly from the output vocabulary.\\
{\bf I-vector speaker adaptation}~\cite{saon13} appends a speaker identity vector to the input features coming from a given speaker. When used in conjunction with speed and tempo perturbation, the perturbed audio recordings of a speaker are considered to be different speakers for the purpose of universal background model (UBM) training, total variability matrix training, and i-vector extraction.\\
{\bf Alignment-length synchronous decoding}~\cite{saon20} is a beam search technique with the property that all competing hypotheses within the beam have the same alignment length. It has been shown to be faster than time-synchronous search for the same accuracy.\\
{\bf Density ratio (DR) LM fusion}~\cite{mcdermott19} is a shallow fusion technique that combines two language models: an external LM trained on a target domain corpus and a language model trained on the acoustic transcripts (source domain) only. The latter is used to subtract the effect of the intrinsic LM given by the prediction network (idea further developed in~\cite{variani20}). Decoding using DR fusion is done according to:

\begin{equation}
{\bf y}^* = \argmax_{{\bf y}\in{\cal Y}^*} \{\log p({\bf y}|{\bf x}) -\mu\log p^{src}({\bf y}) + \lambda\log p^{ext}({\bf y})+ \rho|{\bf y}|\} 
\end{equation}
~~\\
where $\mu,\lambda,\rho$ are the weights corresponding to the source LM $p^{src}$, external LM $p^{ext}$, and label length reward $|{\bf y}|$.

\section{Experiments and results}
\label{experiments}
We investigate the effectiveness of the proposed techniques on one
public corpus (English conversational telephone speech 300 hours) and
two internal datasets (Spanish and Italian conversational speech 780 hours and
900 hours, respectively).

\subsection{Experiments on Switchboard 300 hours}
\label{swb-recipe}
The first set of experiments was conducted on the Switchboard speech
corpus which contains 300 hours of English telephone conversations
between two strangers on a preassigned topic. The acoustic data
segmentation and transcript preparation are done according to the
Kaldi {\tt s5c} recipe~\cite{povey11}. We report results on the
commonly used Hub5 2000 Switchboard and CallHome test sets as well as
Hub5 2001 and RT'03, which are processed according to the LDC
segmentation and scored using Kaldi scoring for measuring WER.

We extract 40-dimensional speaker independent log-Mel filterbank
features every 10 ms which are mean and variance normalized per
conversation side. The features are augmented with
$\Delta$ and $\Delta\Delta$ coefficients and every
two consecutive frames are stacked and every second frame is skipped
resulting in 240-dimensional vectors extracted every 20 ms. These
features are augmented with 100-dimensional i-vectors that are
extracted using a 2048 40-dimensional diagonal covariance Gaussian
mixture UBM trained on speaker independent PLP features transformed
with LDA and a semi-tied covariance transform.

Speed and tempo perturbation is applied to each conversation side with
values in $\{0.9,1.1\}$ for both speed and tempo separately resulting
in 4 additional training data replicas which, together with the
original data, amounts to 1500 hours of training data per epoch. For
sequence noise injection, we add, with probability 0.8, to the
spectrum of each training utterance the spectrum of one random
utterance of similar length scaled by a factor of 0.4. For SpecAugment
we used the settings published in~\cite{park19}. Lastly, in label
switchout, for a sequence of length $U$, we first sample
$\hat{n}\in\{0,\ldots,U\}$ with $p(\hat{n})\propto e^{-\hat{n}/\tau}$
and then we replace, with probability $\hat{n}/U$, the true characters 
with random characters for each position in the sequence. We set the
temperature to $\tau=10$ in our experiments.

The architecture of the trained models is as follows. The encoder
contains 6 bidirectional LSTM layers with 640 cells per layer per
direction and is initialized with a network trained with CTC based
on~\cite{audhkhasi19} similar to~\cite{graves13,rao17,wang18,zeyer20}.
The prediction network is a single unidirectional LSTM layer with only
768 cells (this size has been found to be optimal after external LM
fusion). The joint network projects the 1280-dimensional stacked
encoder vectors from the last layer and the 768-dimensional prediction
net embedding to 256 dimensions and combines the projected vectors
using either $+$ or $\odot$. After the application of hyperbolic
tangent, the output is projected to 46 logits followed by a softmax
layer corresponding to 45 characters plus BLANK. All models have 57M
parameters.

\begin{table}[H]
\begin{center}
\begin{tabular}{|l|l|c|c|c|} \hline
Optimizer    & LR policy & Batch size & SWB  & CH \\ \hline
Momentum SGD & const+decay & 256        & 9.6 & 17.5 \\ \hline
Momentum SGD & const+decay & 64        & 9.4 & 17.5 \\ \hline
AdamW        & const+decay & 256        & 9.4 & 17.8 \\ \hline
AdamW        & const+decay & 64        & 8.5 & 16.7 \\ \hline
AdamW        & OneCycleLR & 64     & 8.1 & 16.5 \\ \hline
\end{tabular}
\end{center}
\caption{\label{tab1} Effect of optimizer, batch size and learning rate schedule on recognition performance for Switchboard 300 hours (Hub5'00 test set).}
\end{table}

The models were trained in Pytorch on V100 GPUs for 20 epochs using
SGD variants that differ in optimizer, learning rate schedule and
batch size.  In Table~\ref{tab1} we look at the effect on performance
of changing the learning strategy. For momentum SGD, the learning rate
was set to 0.01 and decays geometrically by a factor of 0.7 every
epoch after epoch 10.  For AdamW, the maximum learning rate is set to
5e-4 and the OneCycleLR policy~\cite{smith19} consists in a linear
warmup phase from 5e-5 to 5e-4 over the first 6 epochs followed by a
linear annealing phase to 0 for the next 14 epochs. As can be seen,
AdamW with OneCycleLR scheduling and a batch size of 64 appears to be
optimal in this experiment.

\begin{table}[H]
\begin{center}
\resizebox{\columnwidth}{!}{\begin{tabular}{|l|c|c|c||c|c|c|} \hline
\multirow{2}{*}{Model} & \multicolumn{3}{|c||}{No ext LM} & \multicolumn{3}{|c|}{With ext LM} \\ \cline{2-7}
               & SWB  & CH  & Avg & SWB  & CH  & Avg\\ \hline
Baseline       & 7.9  & 15.7&  11.8   & 6.4 & 13.4 & 9.9  \\ \hline
No Switchout   & 8.1  & 15.5&  11.8   & 6.3 & 13.1 & 9.7  \\ \hline
No Seq. noise  &8.4& 16.1&  12.2   & 6.6  & 13.8 & 10.2  \\ \hline
No i-vectors   &8.1   & 16.0&  12.0   & 6.6 & 13.9 & 10.3 \\ \hline
No CTC init.   & 8.3&  16.3&  12.3 & 6.6 & 14.1 & 10.4 \\ \hline
No Density ratio & 7.9  & 15.7&  11.8   & 6.9 & 14.4 & 10.7  \\ \hline
No DropConnect &8.2 & 16.7 & 12.5& 7.0 & 14.8 & 10.9 \\ \hline
No SpecAugment &8.8 & 18.1 & 13.5 & 6.9 & 15.5 & 11.2 \\ \hline
No Speed/tempo &10.0& 18.3 & 14.2 & 7.6 & 15.2 & 11.4   \\ \hline
\end{tabular}}
\end{center}
\caption{\label{tab2} Ablation study on Switchboard 300 hours (Hub5'00 test set).}
\end{table}

Next, we perform an ablation study on the final recipe to tease out
the importance of the individual components. We show results in
Table~\ref{tab2} for models with multiplicative integration with and
without external language model fusion. The source LM is a one-layer
LSTM with 768 cells (5M parameters) trained on the Switchboard 300 hours
character-level transcripts (15.4M tokens) whereas the target LM is a
two-layer LSTM with 2048 cells per layer (84M parameters) trained on
the Switchboard+Fisher character-level transcripts (126M tokens).

Surprisingly, deactivating switchout from the final recipe actually
improves recognition performance after external LM fusion, which was
not the case in prior experiments where this technique was marginally
helpful. Also, another unexpected finding was the large gain due to
density ratio LM fusion. We attribute this to optimizing the other
elements of the training recipe around this technique (e.g. reducing
the size of the prediction network). 

\begin{table}[H]
\begin{center}
\resizebox{\columnwidth}{!}{\begin{tabular}{|c|c|c|c||c|c|c||c|c|} \hline
ext & \multirow{2}{*}{Model} & \multicolumn{2}{|c||}{Hub5'00} & \multicolumn{3}{|c||}{Hub5'01}& \multicolumn{2}{|c|}{RT'03}\\ \cline{3-9}
LM  &                        & swb  & ch  & swb & s2p3 & s2p4 & swb & fsh  \\ \hline
\multirow{3}{*}{no}   & $+$  & 8.0  & 15.6& 8.7 & 11.4 & 16.3 & 18.0& 11.4 \\ \cline{2-9}
    &                $\odot$ & 8.1  & 15.5& 8.5 & 11.7 & 15.9 & 18.5& 11.8 \\ \cline{2-9}
    &               Comb.     & 7.5  & 14.3& 7.9 & 10.8 & 15.2 & 17.1& 10.7 \\ \hline
\multirow{3}{*}{yes} & $+$   & 6.4  & 13.4& 7.0 &  9.2 & 13.4 & 15.0& 9.2  \\ \cline{2-9}
    &              $\odot$   & 6.3  & 13.1& 7.1 &  9.4 & 13.6 & 15.4& 9.5  \\ \cline{2-9}
    &               Comb.     &{\bf 5.9}&  {\bf 12.5}&{\bf 6.6} & {\bf 8.6} & {\bf 12.6} & {\bf 14.1} & {\bf 8.6} \\ \hline
\end{tabular}}
\end{center}
\caption{\label{tab3} Recognition results for additive, multiplicative and combined RNN-Ts on Switchboard 300 hours (Hub5'00, Hub5'01, RT'03 test sets).}
\end{table}

In the next experiment, we compare RNN-Ts with additive versus
multiplicative integration in the joint network. Based on the previous
findings, we train these models without switchout. We also perform
log-linear model combination with density ratio LM fusion according
to:

\begin{align}
\nonumber {\bf y}^* = \argmax_{{\bf y}\in  {\cal H}_{+}({\bf x})\cup{\cal H}_{\odot}({\bf x})} \{ & \alpha\log p({\bf y}|{\bf x};\theta_+)+\beta\log p({\bf y}|{\bf x};\theta_\odot)\\ 
& -\mu\log p^{src}({\bf y}) + \lambda\log p^{ext}({\bf y})+ \rho|{\bf y}|\} 
\end{align}
~~\\
where ${\cal H}_{+}({\bf x}), {\cal H}_{\odot}({\bf x})$ are the
n-best hypotheses generated by the additive and multiplicative RNN-Ts.
Concretely, we rescore the union of the top 32 hypotheses from each
model with $\alpha=\beta=\mu=0.5,\lambda=0.7$ and $\rho=0.2$. The
results from Table~\ref{tab3} show that multiplicative RNN-Ts are
comparable on Hub5'00 and Hub5'01 but slightly worse on RT'03 and that model
combination significantly improves the recognition performance across
all test sets.

For comparison, we include in Table~\ref{tab4} the top performing
single model systems from the literature on the Switchboard 300 hours
corpus. The numbers should be compared with 6.3/13.1 (row 5 from
Table~\ref{tab3}). As can be seen, the proposed modeling techniques
exhibit excellent performance on this task.

\begin{table}[H]
\begin{center}
\begin{tabular}{|l|l|c|c|c|}\hline
System & Type & ext. LM & SWB & CH\\ \hline
Park et al.'19~\cite{park19} & Att. enc-dec & LSTM & 6.8 & 14.1\\ \hline
Irie et al.'19~\cite{irie19} & Hybrid & Transf. & 6.7 & 12.9 \\ \hline
Hadian et al.'18~\cite{hadian18} & LF-MMI & RNN & 7.5 & 14.6 \\ \hline
T{\"u}ske et al.'20~\cite{tuske20} & Att. enc-dec & LSTM & 6.4 & 12.5 \\ \hline
\end{tabular}
\end{center}
\caption{\label{tab4} Single model performance for existing systems on Switchboard 300 hours (Hub5'00 test set).}
\end{table}

\subsection{Experiments on conversational Spanish 780 hours}
We also investigated the effectiveness of the proposed techniques on
an internal dataset which consists of 780 hours of Spanish call center
data collected using a balanced distribution of Castilian and other
Central and South American dialects from roughly 4000 speakers.  The
test set on which we report results comes from the call center domain and
contains 4 hours of speech, 113 speakers and 31.6k words.

The model architecture, training and decoding recipes are identical to
the ones from the previous subsection (except for a 40-character output layer) with the difference that we do
not use i-vector speaker adaptation because of the small amount
of data per speaker. The external language model
is a two layer LSTM with 2048 cells per layer (84M parameters)
trained on the character-level acoustic transcripts (36M characters) as well as
additional text data from the customer care domain (173M characters). The
source LM is a single-layer LSTM with 1024 cells (8.7M parameters).

\begin{table}[H]
\begin{center}
\begin{tabular}{|l|c|c|c|c|}\hline
Model/technique & Training data & WER\\ \hline
Initial experiment & 250 h & 34.8\\ \hline
+ DropConnect+seq. noise & 250 h & 27.6 \\ \hline
+ Speed/tempo & 780 h & 25.0\\ \hline
+ CTC encoder pretraining & 780 h & 23.6\\ \hline 
+ Multiplicative integration & 780 h & 22.7\\ \hline
+ SpecAugment & 780 h & 21.9\\ \hline
+ AdamW+OneCycleLR & 780 h & 20.8\\ \hline
+ Shallow LM fusion & 780 h & 20.3\\ \hline
+ Density ratio LM fusion & 780 h & 20.0\\ \hline
\end{tabular}
\end{center}
\caption{\label{tab5} Cumulative effect of proposed techniques on conversational Spanish 780 hours (internal test set).}
\end{table}

In Table~\ref{tab5}, we look at the impact of the various techniques
on word error rate for bidirectional RNN-Ts. The first 4 models have
additive integration and the first 6 models are trained with momentum
SGD and a constant+decay learning rate schedule for 20 epochs. Unlike
in the previous experiments, here we observe a significant gain from
multiplicative integration (0.9\% absolute) and a smaller gain from density ratio fusion (0.3\% absolute).

\subsection{Experiments on conversational Italian 900 hours}

The last set of experiments was conducted on an internal Italian
corpus of conversational speech that was collected using scripted
dialogs from several domains such as banking, insurance, telco, retail
to name a few. The data has a balanced demographic and dialectal distribution
with a large fraction of the speakers speaking standard dialect
(as spoken in the capital or news broadcasts). We report results on the
Mozilla CommonVoice\footnote{\tt https://commonvoice.mozilla.org} Italian test set which has 1.2 hours
of audio, 764 utterances and 6.6K reference words.

Similar to the Spanish setup, we trained character-level RNN-Ts (53
outputs) without i-vector speaker adaptation using the recipe
described in section~\ref{recipe} with specifics from
subsection~\ref{swb-recipe}.  In addition, we also train RNN-Ts with
unidirectional LSTM encoders with 6 layers and 1024 cells per layer on
stacked log-Mel features augmented with $\Delta,\Delta\Delta$ with 5
frames lookahead as proposed in~\cite{li19}. The language model
configuration for density ratio fusion is as follows: the source LM
has one LSTM layer with 1024 units (8.7M parameters) and is trained on
the character-level acoustic transcripts (41.7M characters) whereas
the external LM is a two layer 1024 cells/layer LSTM (21.3M
parameters) trained on the acoustic transcripts and 10\% of Italian
Wikipedia data (306M characters).

In Table~\ref{tab6}, we compare additive versus multiplicative models
with bidirectional and unidirectional encoders with and without
external LM fusion (regular and density ratio). Based on these
results, three observations can be made. First, multiplicative
integration significantly outperforms the additive counterpart for
unidirectional models and after shallow LM fusion (for both).  Second,
there is a severe degradation in recognition performance from using
unidirectional encoders which can be mitigated with techniques
from~\cite{jain19,kurata20}. Third, density ratio LM fusion
significantly outperforms regular shallow fusion which we attribute to
the mismatched training and testing conditions.

\begin{table}[H]
\begin{center}
\begin{tabular}{|l|c|c|c|c|} \hline
                   & \multicolumn{2}{|c|}{Bidirectional} & \multicolumn{2}{|c|}{Unidirectional} \\ \cline{2-5}
                   & $+$ & $\odot$   & $+$  & $\odot$   \\ \hline
No external LM         & 17.8  & 17.6  & 28.0 & 26.2 \\ \hline
Shallow fusion     & 15.2  & 13.9  & 22.9 & 21.4 \\ \hline
Density ratio fusion & 13.6  & 12.7  & 20.9 & 18.6 \\ \hline
\end{tabular}
\end{center}
\caption{\label{tab6} Various comparisons on conversational Italian 900 hours (Mozilla CommonVoice test set).}
\end{table}

\section{Conclusion}
\label{conclusion}
The contribution of this paper is twofold. First, we have introduced a
simple yet effective modification of the joint network whereby we
combine the encoder and prediction network embeddings using
multiplicative integration instead of additive. MI outperforms
additive integration on two out of three tasks and shows good model
complementarity on the Switchboard 300 hours corpus. Second, we have
shown that careful choice of optimizer and learning rate scheduling,
data augmentation/perturbation and model regularization, speaker
adaptation, external language model fusion, and model combination can
lead to excellent recognition performance when applied to models with
a very simple architecture and character outputs. Future work will
look at alternative neural transducer architectures
(e.g.~\cite{yeh19}) and training criteria (e.g.~\cite{weng19}) and attempt to
simplify the training recipe without sacrificing recognition
performance.

\bibliographystyle{IEEEbib}
\bibliography{icassp2021}
\end{document}